\pdfoutput=1

\documentclass[11pt]{article}

\usepackage[preprint]{acl}

\usepackage{times}
\usepackage{latexsym}
\usepackage{geometry}
\usepackage{graphicx} 
\usepackage{booktabs} 
\usepackage{makecell}
\usepackage{tabularx}
\usepackage{caption}
\usepackage{subcaption}
\usepackage{todonotes}
\usepackage{multirow}
\usepackage{arydshln} 

\usepackage[T1]{fontenc}

\usepackage[utf8]{inputenc}

\usepackage{microtype}

\usepackage{inconsolata}

\usepackage{graphicx}

\title{Quality or Quantity? On Data Scale and Diversity in Adapting \\Large Language Models for Low-Resource Translation}

\author{Vivek Iyer\quad Bhavitvya Malik\thanks{denotes equal contribution}\quad Pavel Stepachev\footnotemark[1]\quad \\
  \bf{Pinzhen Chen\quad Barry Haddow\quad Alexandra Birch} \\
  School of Informatics, University of Edinburgh \\
  \texttt{vivek.iyer@ed.ac.uk} \\}

\begin{document}
\maketitle
\begin{abstract}
Despite the recent popularity of Large Language Models (LLMs) in Machine Translation (MT), their performance in low-resource languages (LRLs) still lags significantly behind Neural Machine Translation (NMT) models. In this work, we explore what it would take to adapt LLMs for the low-resource setting. Particularly, we re-examine the role of two factors: a) the importance and application of parallel data, and b) diversity in Supervised Fine-Tuning (SFT). Recently, parallel data has seen reduced use in adapting LLMs for MT, while data diversity has been embraced to promote transfer across languages and tasks. However, for low-resource LLM-MT, we show that the opposite is true for both considerations: a) \emph{parallel data} is critical during both pre-training and SFT; b) diversity tends to cause \emph{interference} instead of transfer. Our experiments with three LLMs across two low-resourced language groups---Indigenous American and North-East Indian---reveal consistent trends, underscoring the generalizability of our findings. We believe these insights will be valuable for scaling to massively multilingual LLM-MT models that can effectively serve LRLs.
\end{abstract}

\section{Introduction}
\label{sec:introduction}
Large Language Models (LLMs) have been dominating recent research in Machine Translation (MT), showing good few-shot prompting \citep{garcia2023unreasonable, hendy2023good} and stronger instruction-tuning \citep{alves2024tower, xu2024a} performances---recently even outperforming commercial Neural Machine Translation (NMT) models \citep{kocmi2024preliminarywmt24rankinggeneral}. 
However, LLM translation for low-resource languages (LRLs) still lags significantly behind NMT models \citep{robinson2023chatgpt, zhu-etal-2024-multilingual}. While the strong performance of LLMs on high-resource languages can be attributed to the skewed language distribution during pre-training and the unintentional consumption of parallel data at scale \citep{briakou-etal-2023-searching}, no such relief exists for LRLs.
This leads to the main question motivating this paper: \textit{What would it take to adapt LLMs for low-resource MT}? 

Recent work on LRL translation with LLMs has explored using resources like multilingual lexicons \citep{lu2023chain}, word alignments \citep{mao2024tuning} or linguistic tools \citep{zhang2024hire}. While effective, reliance on such tools hinders ease of extensibility across languages. Instead, in this work, we take inspiration from research done for high-resource translation with LLMs, where a 2-stage training paradigm of Continued Pre-Training (CPT), followed by small-scale Supervised Fine-Tuning (SFT; \citep{xu2024a, alves2024tower}) has been successful. Aiming to adapt this framework for low-resource MT, we re-examine the role of two factors influencing the performance of translation LLMs: a) \textit{how best to leverage parallel data}, and b) \textit{the interplay between diversity and transfer} during SFT (also known as `instruction tuning').

\begin{table}[t]
\centering\small\setlength{\tabcolsep}{0.5ex}
\begin{tabular}{llccc}
\toprule
& \multicolumn{1}{c}{\textbf{Base LLM}} & \textbf{\#Tokens} & \textbf{\#Langs} \\
\midrule
\citealp{zhang2024when} & From scratch & 283B & \phantom{10}2 \\
\citealp{fujii2024continual} & Llama2 & 100B & \phantom{10}2 \\
\citealp{lu2024llamax} & Llama\{2,3\} & \phantom{$^\dag$0}$\sim$82B$^\dag$\phantom{$\sim$} & 101 \\  
\citealp{alves2024tower} & Llama2 & \phantom{0}20B & \phantom{0}11 \\
\citealp{xu2024a} & Llama2 & \phantom{0}20B & \phantom{10}6 \\
Ours & Mistral/Llama3 & \phantom{0}0.7B\phantom{.} & \phantom{0}12 \\
\bottomrule
\end{tabular}
\caption{Comparing data scales of previous works with ours, in terms of pre-training token counts, the base LLM (if pre-training continued from one) and how many languages this spanned. $^\dag$Estimated from the reported sentence count assuming 100 tokens per sentence.}\vspace{-2ex}
\label{table:scale_comparison_relatedworks}
\end{table}


Recently, the role of \textbf{parallel data at scale}, long viewed as fundamental to the success of NMT models, has come into question in the era of LLM-MT systems.
Motivated by the modest gains of training on 300M parallel sentences \citep{yang2023bigtranslate}, and the surprising benefits of scaling \emph{down} during SFT \citep{zhou2023lima}, subsequent works have used only tens of thousands of human-written bitext for LLM-MT \citep{zhang2023bayling, alves2024tower, xu2024a}---with SFT scaling laws further showing the early plateau of LLM-MT performance \citep{zhang2024when}. Even more surprisingly, 
\citet{zhu2024fine} showed MT abilities emerging with just 32 SFT examples! However, these explorations concern LLMs pre-trained on several billions of tokens in the languages in question. We revisit these notions for low-resource MT and work with languages having datasets that are 2-3 orders of magnitude smaller. In Table \ref{table:scale_comparison_relatedworks} we compare the scale of the datasets used in our work and related research. We discover that for low-resource MT, \textit{parallel data is critical} not just during CPT, but even more so during SFT---in direct contrast with research on high-resource languages (HRLs). 

Next, \textbf{diversity} in tasks, prompts, and datasets during SFT has been shown to significantly improve model performance across a range of tasks \citep{mishra-etal-2022-cross, chung2024scaling}. MT instructions have been shown to not just boost translation performance in unseen languages~\citep{muennighoff-etal-2023-crosslingual}, but also enhance LLM capabilities across diverse multilingual generation tasks \citep{ranaldi-pucci-2023-english, zhu2023extrapolating}. Inspired by these findings, we study if SFT diversity could benefit low-resource LLM-MT systems too. By conducting experiments across a range of tasks and language pairs with SFT datasets of varying compositions, we establish that \textit{diversity leads to negative interference} and fine-tuning on multilingual MT is the optimal strategy. Further, we observe that training for more epochs on MT data is more effective than curating and training on a diverse dataset of the same size.

Our contributions are thus as follows:

\begin{enumerate}
    \item In contrast to  findings for high-resource LLM-MT \citep{xu2024a}, we observe that for LRLs, LLMs benefit hugely from scale of parallel data, during both CPT and SFT stages
    \item Linguistic and task diversity during SFT leads to negative interference for LRL LLM-MT, with focused multilingual MT fine-tuning for more epochs being the most effective recipe.
\end{enumerate}

 To ensure the generalizability of our findings, we conduct 2 sets of experiments training multilingual LLMs on different sets of languages: a) 11 Indigenous American and b) 4 North East Indian languages, wherein the former follows a Latin script and the latter includes languages that do not. 
Our focus in this work is on the eng/spa $\rightarrow$ X directions since generation in an LRL is known to be a much harder task than in an HRL like English or Spanish, and we are interested in studying the challenges involved. We experiment with 3 base LLMs of varying sizes---Gemma 2B \citep{team2024gemma}, Mistral 7B \citep{jiang2023mistral}, and Llama 3 8B \citep{dubey2024llama}, and observe that findings are mostly consistent across these models.

By applying our findings to 2-stage training, our methods achieved a +16.5 average chrF++ improvement over few-shot prompting---with the largest gains coming from the 8 least-resourced American languages in our setup, all of which have about 10K-50K parallel sentences each. We hope that the findings of this work will be useful when scaling to LLMs that can effectively translate into lower-resource languages.





\section{Related Work}
\label{sec:relatedwork}
\paragraph{High-Resource Translation with LLMs} There has been considerable interest in using LLMs as MT systems recently. Following initial success in prompting LLMs for high-resourced pairs \citep{vilar-etal-2023-prompting, garcia2023unreasonable, hendy2023good, zhang2023prompting, iyer2023towards}, subsequent works have attempted to train LLMs on parallel data at scale \citep{yang2023bigtranslate, lu2024llamax}, but these yielded modest gains and underperformed smaller encoder-decoder baselines such as NLLB-200 \citep{costa2022no}. \citet{zhang2024when} showed through scaling laws for SFT that LLMs pre-trained at the order of 50B-300B tokens saturate in MT performance with 20K-30K instructions. Concurrently, \citet{xu2024a} discovered excess parallel data washed out LLM knowledge, so they proposed a 2-stage paradigm called ALMA that involved pre-training on scaled-up monolingual data, followed by SFT on much smaller high-quality bitext (\textasciitilde 60K lines). ALMA outperformed NLLB-200.  Following their success, \citet{alves2024tower} adopted the ALMA framework to train Tower 7B for 10 high-resourced languages, outperforming ALMA and also matching GPT-4.

It is worth noting that \citet{xu2024a} did not include parallel data during CPT. However, inspired by research showing LLMs unintentionally consume parallel data at scale \citep{briakou-etal-2023-searching}, several works have included it at the order of several billions of tokens for the top 10-20 high-resource languages \citep{anil2023palm, wei2023polylm, fujii2024continual, alves2024tower}. Concurrent work has explored pre-training on synthetic, code-switched parallel data for 101 languages \citep{lu2024llamax} with a total of 400M sentences. 
This work explores the impact of parallel data exclusively for low-resource performance and experiments with 1M--13M parallel sentences (50M--750M tokens) during pre-training, two orders of magnitude smaller than prior work. 

\paragraph{Low-Resource Explorations in LLM-MT} LLMs have been shown to perform poorly in low-resource MT \citep{robinson2023chatgpt, zhu-etal-2024-multilingual}. In response, there have been efforts to leverage external resources in the MT pipeline, including multilingual lexicons \citep{lu2023chain}, rule-based linguistic tools \citep{zhang2024hire}, word alignments \citep{mao2024tuning} and even entire grammar books \citep{reid2024gemini}. However, such approaches create dependencies on resources and hinder extensibility across languages. Instead, we focus on optimal data utilisation strategies during CPT and SFT, prioritizing extensibility.

\paragraph{Cross-Lingual Instruction Tuning} There has been a body of work exploring multilingual instruction tuning that have touched on diversity and data scale, but most of it is limited to HRLs. MT examples were shown to improve cross-lingual generation \citep{ranaldi-pucci-2023-english, zhu2023extrapolating}, while \citet{chen-etal-2024-monolingual} showed multilingual SFT on machine-translated Alpaca datasets matches or beats monolingual tuning. \citet{kew2023turning} and \citet{shaham2024multilingual} showed a small quantity of multilingual SFT data can improve cross-lingual generation capabilities in medium and high-resource languages, while \citet{zhu2024fine} showed only 32 examples in HRLs suffice to elicit MT capabilities from LLMs. In our work, we hypothesize that these findings have the common denominator of pre-training on HRLs at scale, and show that when moving to LRLs, these trends reverse and diversity is no longer beneficial.





\section{Approach}
\label{sec:approach}
We now describe our efforts to adapt the widely-used ALMA framework, originally designed for fine-tuning LLMs for HRL translation \citep{xu2024a, alves2024tower}, for MT in LRLs.

\subsection{Stage 1: Continued Pre-training (CPT)}
\label{sec:continued-pt}

\paragraph{CPT on Monolingual Data} The objective of this stage is to `teach' an LLM to model LRLs, which are scarce in the pre-training corpus. We conduct CPT on monolingual data with the standard Causal Language Modelling objective. We train with low-rank adaptation (LoRA) and attach rank 8 adapters to query and value matrices \citep{hu2022lora}. We also fine-tune input and output embeddings.

\paragraph{CPT on Parallel Data} In a scenario where monolingual data is scarce, it is crucial to investigate the most effective way to use parallel data---which, for our indigenous American languages, was found to surprisingly be more abundant than the former\footnote{While monolingual online data is scarce, many translations of constitutions, articles etc. from Spanish do exist}. We investigate 3 methods of mixing all available parallel and monolingual data:
\begin{enumerate}
    \item \textbf{All Mono}: Here, we merge monolingual data with only \textit{the target side} of all available bitext---essentially using it as extra monolingual data
    \item \textbf{Mono + parallel (concat)}: Here, we merge monolingual data with \emph{concatenated source-target pairs} from parallel data.  We prepend source and target language codes before concatenation and, following \citet{guo2024novel}, use a newline delimiter to separate them.
    \item \textbf{Mono + parallel (separate)}: To ablate the impact of concatenation, we provide the source and target sides of parallel data as \emph{separate} sentences, and shuffle with monolingual data.
\end{enumerate}

\begin{figure}[t]
    \centering
    \includegraphics[clip,trim=4.5cm 4.5cm 4.5cm 4.5cm ,width=\columnwidth]{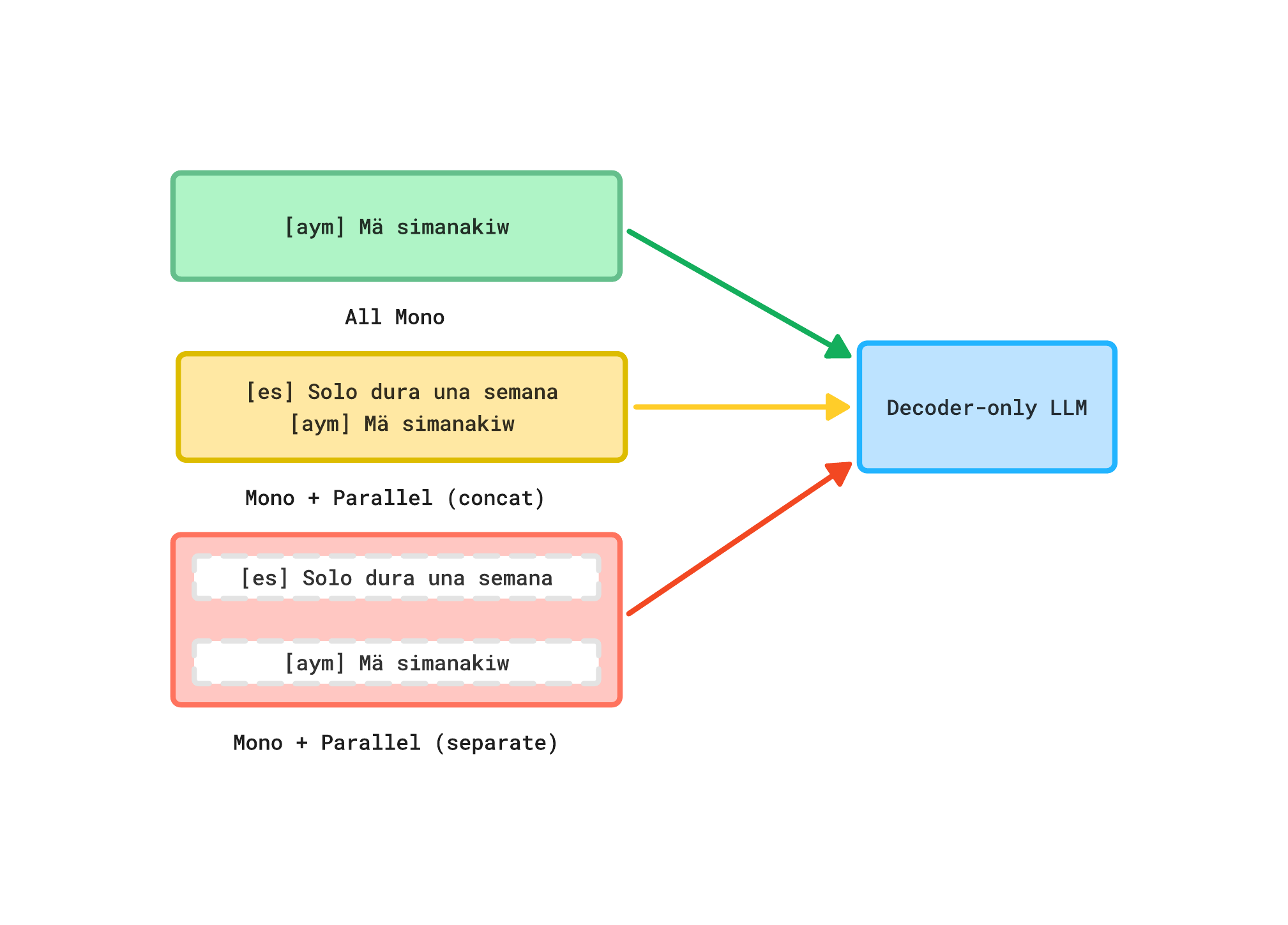}
    \caption{Strategies explored for incorporating parallel data during Continued Pre-Training. We show a Spanish (\texttt{es}) to Aymara (\texttt{aym}) example from our parallel data.}\vspace{-2ex}
    \label{fig:architecture}
\end{figure}

 We depict our approach in Figure \ref{fig:architecture}. In the first technique, the motivation is that it might be hard for the LLM to learn to model concatenated sequences, given that they were likely scarce in the original pre-training corpus, with the added challenge of pre-training on `new' low-resource languages. 
On the other hand, if the model is able to adapt to concatenated sequences, it could make the LLM more adjusted to the task of translation. 
Finally, the third method verifies whether the results of the `concatenated' model are due to the concatenation itself, rather than simply being exposed to additional tokens in the source language. We use this terminology for all experiments in this work.

While these methods control how parallel data is incorporated in pre-training, we are also interested in adjusting the ratio of parallel to monolingual data in the corpus, particularly for the \textit{`concat'} method given pre-training on 100\% parallel data can be suboptimal \citep{alves2024tower}. To get a desired mixing ratio for \textit{`concat'}, we include bitext only until it comprises a given percentage of the training corpus. Once this threshold is crossed, we use the \textit{`all mono'} method to include it as monolingual data instead. Next, we  use temperature sampling \citep{arivazhagan2019massively} to control the language-wise distribution in our monolingual and parallel pre-training data, since these are quite heterogeneous and certain languages are extremely low-resourced. We sample monolingual and parallel data independently if using the `concat' method, else we just mix them all together and shuffle at the instance level to create our final pre-training corpus.

\subsection{Stage 2: Supervised Fine-Tuning (SFT)}
\label{sec:instruction-tuning}

Next, we fine-tune with LoRA on supervised instruction data and detail the tasks explored below. Note that we convert all instructions to the standard Alpaca format, and compute loss on the target tokens only \citep{alpaca}. 

\paragraph{Low-Resource MT} Given our use case, the most intuitive task to include would be MT itself. The \texttt{instruction} for each example is chosen randomly from a set of translation prompts (Table \ref{tab:MTInstructionTemplates}), while the \texttt{input} and \texttt{output} fields are the source and target sentences respectively. 

\paragraph{High-Resource MT} Apart from MT data in the LRLs, we also experiment with adding HRL MT data since it is more abundant and of higher quality, known to be important during SFT \citep{xu2024a}. To explore the impact of transfer learning, we work with HRL data that is in some way related to the source/target language, e.g.\ Spanish-English data for experiments on Spanish-X. Instructions are formatted in the same way as the LRL data.

\paragraph{General-Purpose Instruction Tuning} Apart from MT data, we also explore adding widely used general-purpose instruction tuning datasets, such as Alpaca \citep{alpaca} and Aya \citep{singh2024aya}, and use data from high-resource languages (comprising the source side in our tasks) to improve the model's overall instruction-following capabilities. However, for the most part, we are unable to find similar data in the LRLs we experiment on.

\paragraph{Synthetic Cross-Lingual QA (XQA)} We do not find any instruction tuning data for most LRLs, so we follow \citet{iyer-etal-2024-exploring} to create synthetic Question Answering (QA) data. Starting from a parallel sentence pair $(X,Y)$, where $X$ is from an HRL (in our case, English/Spanish) and $Y$ is from an LRL, we prompt an LLM (Mixtral-8x7B-Instruct~\cite{jiang2024mixtral} in this work) to generate a question $Q$ for which $X$ would be an answer. Since $X$ and $Y$ are semantically equivalent, $Y$ is treated as the answer to question $Q$. 
We add a requirement at the end of $Q$ to generate in the target language. Thus, we use $(Q, Y)$ as synthetic cross-lingual instruction data. We provide the templates used for generating XQA examples in Table \ref{tab:XQAInstructionTemplates}.



\section{Experiments and Discussions}
\label{sec:experiments}
We fine-tune 2 separate sets of multilingual LLMs for 2 different language groups to facilitate evaluation on test sets from 2 different low-resource MT shared tasks: AmericasNLP 2024 \citep{ebrahimi2023findings} and the Indic track of WMT 2023 \citep{pal2023findings}. The former involves 11 Indigenous Central \& South American languages, while the latter focuses on 4 North-East (NE) Indian languages. The first group includes Aymara (\texttt{aym}), Bribri (\texttt{bzd}), Asháninka (\texttt{cni}), Chatino (\texttt{ctp}), Guarani (\texttt{grn}), Huichol (\texttt{hch}), Nahuatl (\texttt{nhe}), Otomi (\texttt{ote}), Quechua (\texttt{quy}), Shipibo-Konibo (\texttt{shp}) and Tarahumara (\texttt{tar}). The second consists of Khasi (\texttt{kha}), Meitei (\texttt{mni}), Mizo (\texttt{lus}) and Assamese (\texttt{asm}). Our motivation in choosing these languages was to experiment with LRLs containing both Latin (American) and non-Latin (Indic) scripts; that also had widely used, high-quality test sets. We use the former for our main experiments and replicate the most interesting baselines in the Indic languages.

\subsection{Data}
\begin{table}[t]
\centering\small
\setlength{\tabcolsep}{1ex}
\begin{subtable}{\linewidth}
\centering\small
\begin{tabular}{lclclc}
\toprule
\textbf{Lang} & \textbf{\#Tokens} & \textbf{Lang} & \textbf{\#Tokens} & \textbf{Lang} & \textbf{\#Tokens} \\
\midrule
aym   & 23.4M    & bzd  & \phantom{0}2.6M    & cni  & \phantom{00}2.2M    \\
ctp  & \phantom{0}5.4M    & grn   & 37.6M   & hch  & \phantom{00}3.5M    \\
nhe  & 32.1M    & oto  & 23.6M   & quy  & \phantom{0}45.1M   \\
shp  & \phantom{0}3.3M     & tar  & \phantom{0}2.3M    & \textbf{Total} & \textbf{181.3M} \\
\midrule
eng & \phantom{0}9.8M      & spa   & 27.9M & \textbf{Replay} & \phantom{0}\textbf{37.6M} \\
\bottomrule
\end{tabular}
\caption{Indigenous American Languages}
\end{subtable}

\vspace{0.5em} 

\begin{subtable}{\linewidth}
\centering\small
\begin{tabular}{lclclc}
\toprule
\textbf{Lang} & \textbf{\#Tokens} & \textbf{Lang} & \textbf{\#Tokens} & \textbf{Lang} & \textbf{\#Tokens} \\
\midrule
asm   & \phantom{0}1.1B    & kha  & 39.3M   & lus  & 165.1M  \\
mni   & 16.5M   & eng  & \phantom{0}7.3M    & \textbf{Total} & \phantom{00}\textbf{1.3B} \\
\bottomrule
\end{tabular}
\caption{North-East Indian Languages}
\end{subtable}

\caption{Monolingual data statistics, with token counts calculated using the Llama3 8B tokenizer. English and Spanish are included as replay data. Note that LRL token counts overestimate data sizes, due to poor tokenization, and cannot be directly compared with HRLs.}
\label{tab:mono_data_Stats}
\end{table}

\begin{table*}[h]
\centering\small
\setlength{\tabcolsep}{1ex}
\begin{tabular}{l@{\hskip 3ex}ccc@{\hskip 3ex}ccccccccccc}
\toprule
 & \textbf{Total} & \textbf{HRL} & \textbf{LRL} & \textbf{aym} & \textbf{bzd} & \textbf{cni} & \textbf{ctp} & \textbf{grn} & \textbf{hch} & \textbf{nhe} & \textbf{oto} & \textbf{quy} & \textbf{shp} & \textbf{tar} \\
\cmidrule(r){1-4}\cmidrule(lr){5-15}
spa & \phantom{.0}1M & 0.8M & 0.2M & \phantom{.0}442K & 8K & \phantom{0.}20K & 4K & \phantom{00}80K & 21K & 57K & 16K & \phantom{0}226K & 62K & 28K \\
por & 1.9M & 1.9M & \phantom{.}15K & \phantom{00}3.6K & 0 & \phantom{0}6.9K & 0 & \phantom{0}410K & 0 & \phantom{0}8K & 0 & 1520K & 0 & 0 \\
eng & 5.8M & 5.8M & \phantom{.}36K & \phantom{.}1053K & 0 & 13.9K & 0 & 2489K & 0 & 22K & 0 & 2271K & 0 & 0 \\
\midrule
\textbf{Total} & 8.8M & 8.5M & 0.3M & 1.5M & 7.7K & 41K & 4.1K & 3M & 21K & 87K & 16K & 4.0M & 62K & 28K \\
\bottomrule
\end{tabular}
\caption{Parallel data sentence counts for American languages, from source to each target language. \texttt{HRL} and \texttt{LRL} refer to the 3 high-resource languages (Aymara, Guarani, Quechua) and the other 8 low-resource ones respectively.}
\label{tab:american_parallel_stats}
\end{table*}


\begin{table}[h]
\centering\small
\begin{tabular}{@{}lcccc@{}}
\toprule
\textbf{Pair} & \textbf{eng-lus} & \textbf{eng-asm} & \textbf{eng-kha} & \textbf{eng-mni} \\ 
\midrule
\textbf{\#Sents} & 6.5M & 5.0M & 25K & 443K \\ 
\bottomrule
\end{tabular}
\caption{Indic Parallel Data Sizes (Sentence counts)}
\label{tab:indic_parallel_stats}
\end{table}

\paragraph{Monolingual Data} Table \ref{tab:mono_data_Stats}  shows the token counts for the monolingual data collected for the 2 language groups. We note that the American languages are very low-resource, with 6 of 11 having 5M tokens or less. The Indic languages have relatively more data, with Assamese being medium-resourced but still likely low-resource in the original LLM pre-training corpus. Assamese and Meitei follow the Assamese-Bengali script, while Mizo and Khasi use the Latin script. Using the Llama3 tokenizer, we observe average fertilities of 2.87 and 3.83 for the American and Indic languages respectively, almost 3x that of high-resource languages, illustrating the under-representation of non-Latin scripts in SOTA LLMs. Finally, we include some data in English and Spanish as \emph{replay data} to prevent catastrophic forgetting \citep{ibrahim2024simple}. 

\paragraph{Parallel data} We curate parallel data from various sources, for use in both CPT and SFT. Tables \ref{tab:american_parallel_stats} and \ref{tab:indic_parallel_stats} show the sizes for the American and Indian languages respectively. Given that our primary exploration is for the American languages with limited spa-X data, we also sample eng-X and por-X from OPUS \citep{tiedemann2012parallel}. Note the heavily skewed language distribution, with the 3 HRLs constituting ~80\% of spa-X data and 96\% of the overall data. The lesser skew for spa-X is due to the efforts of AmericasNLP to collect data for these pairs and the prevalence of Spanish in Latin American countries. A similar skew also exists for the Indic languages, with English-Khasi being the least-resourced pair. We list sources for all curated data, along with cleaning steps, in Appendix \ref{sec:appendix}.


\begin{table}[t]
    \centering\small\setlength{\tabcolsep}{0.05ex}
    \begin{tabular}{@{}lclclclc@{}}
        \toprule
        \multicolumn{6}{c}{\textbf{Indigenous American}} & \multicolumn{2}{c}{\textbf{NE Indic}} \\
        \cmidrule(lr){1-6} \cmidrule(lr){7-8}
        \multicolumn{1}{c}{\textbf{Pair}} & \textbf{\#Lines} & \multicolumn{1}{c}{\textbf{Pair}} & \textbf{\#Lines} & \multicolumn{1}{c}{\textbf{Pair}} & \textbf{\#Lines} & \multicolumn{1}{c}{\textbf{Pair}} & \textbf{\#Lines} \\
        \midrule
        spa-aym & 996 & spa-nhe & 672 & spa-quy & 996 & eng-asm & 2000 \\
        spa-bzd & 996 & spa-oto & 599 & spa-shp & 996 & eng-kha & 1000 \\
        spa-cni & 883 & spa-gn & 995 & spa-tar & 995 & eng-mni & 1000 \\
        spa-ctp & 499 & spa-hch & 994 & & & eng-lus & 2000 \\
        \bottomrule
    \end{tabular}
    \caption{Evaluation data statistics for the low-resourced American and Indic language experiments in this paper. }
    \label{tab:language_pairs}
\end{table}
\subsection{Evaluation} 
\label{sec:eval}
To evaluate MT into the 11 American languages, we use AmericasNLP'23 validation sets \citep{ebrahimi2023findings} containing spa-X translation pairs. For the Indic pairs, we use the WMT 2023 test sets from the Indic track \citep{pal2023findings} which consist of eng-X pairs. Both evaluation datasets are multi-domain, as are the curated monolingual and parallel corpora. We show test set statistics in Table \ref{tab:language_pairs}. Given the absence of neural metrics for these languages, we evaluate using  ChrF++ \citep{popovic2017chrf++}, since both the American and Indic languages are morphologically rich wherein chrF++ is particularly effective \citep{popovic2017chrf++}. We use SacreBLEU \citep{post-2018-call} for computing this. We also report confidence intervals with bootstrap resampling \citep{koehn2004statistical}, which we implement for the multilingual setting by computing the macro-average across all languages for each resample, and then computing the mean and variance across all resamples. We report the standard deviation as the confidence interval.


\subsection{Experimental Settings}
For temperature sampling our data, we use $\tau=30$ for CPT and $\tau=80$ for SFT. We used a batch size of 8 and gradient accumulation every 16 steps. We used a learning rate of 1e-4, with a cosine scheduler and a warmup ratio of 3\%. We train all models on bf16 precision for 1 epoch. We use Llama-Factory \citep{zheng2024llamafactory} for training and evaluating all models, with Deepspeed ZeRO3 \citep{deepspeed} for distributed training. For inference, we used a batch size of 16 with greedy decoding, since we found higher beam sizes yielded minimal gains.


\begin{table}[h]
    \centering\small
    \setlength{\tabcolsep}{0.5ex}
    \begin{tabular}{lccc}
        \toprule
        \textbf{Fine-Tuned Modules} & \textbf{Gemma\hspace{0.3ex}2B} & \textbf{Mistral\hspace{0.3ex}7B} & \textbf{Llama3\hspace{0.3ex}8B} \\
        \midrule
        LoRA only & 4.5 ± 0.1 & \phantom{1}8.8 ± 0.2 & \phantom{1}8.3 ± 0.2 \\
        LoRA + embeddings & 9.6 ± 0.3 & 15.6 ± 0.4 & 15.6 ± 0.4 \\
        \bottomrule
    \end{tabular}
    \caption{Impact of fine-tuning input/output embeddings along with LoRA adapters is shown. Both models follow the ``CPT \textit{all mono}, SFT'' recipe from Table \ref{tab:primary-results}.}
    \label{tab:embsftablation}
\end{table}

\begin{table}[h]
    \centering
    \setlength{\tabcolsep}{0.2ex}
    \resizebox{0.99\linewidth}{!}{
    \begin{tabular}{lccc}
        \toprule
        \textbf{Method} & \textbf{Gemma 2B} & \textbf{Mistral 7B} & \textbf{Llama3 8B} \\
        \midrule
        5-shot prompting & \phantom{1}2.8 ± 0.1 & \phantom{1}5.1 ± 0.1 & \phantom{1}3.9 ± 0.1\\
        SFT only & \phantom{1}8.7 ± 0.2 & 16.3 ± 0.4 & 14.8 ± 0.4 \\
        CPT \textit{all mono}, SFT & \phantom{1}9.6 ± 0.3 & 15.6 ± 0.4 & 15.6 ± 0.4 \\
        CPT \textit{mono+parallel}, SFT & \textbf{10.1 ± 0.3} & \textbf{16.7 ± 0.4} & \textbf{17.2 ± 0.4} \\
        \bottomrule
    \end{tabular}}
    \caption{chrF++ scores for spa-X LLM-MT in the American languages, for LLMs of varying sizes. Both LoRA modules and embeddings are fine-tuned. For SFT, all models use 500K spa-X MT examples. Confidence estimates are computed using bootstrap resampling. \textit{`mono+parallel'} uses concatenated bitext.}
    \label{tab:primary-results}
\end{table}

\subsection{Foundational Results}
\label{sec:foundational-results}
We first establish the importance of fine-tuning input and output embeddings in Table \ref{tab:embsftablation} and then show our foundational results for the American languages in Table \ref{tab:primary-results}, using 5-shot prompting as a baseline. Our findings are:

\begin{enumerate}

    \item \textbf{Fine-tuning embeddings is critical.} Across the board, we observe that fine-tuning embeddings along with LoRA modules yields huge gains (Table \ref{tab:embsftablation}), almost doubling chrF++ scores, indicating that this step is crucial to helping LLMs adapt to these new languages. Given this, we expect that full-weight fine-tuning would perform better, but stick to LoRA fine-tuning for cost-efficiency reasons. We fine-tune embeddings for all future results.
    
    \item \textbf{Choosing larger base LLMs is crucial.} For under-represented (zero-resource) languages, is it more effective to train smaller LLMs with larger vocabularies (and thus, improved fertility), or vice versa?\footnote{We found it prohibitively expensive to fine-tune Gemma 7B which has a larger vocabulary \emph{and} a larger capacity.} We note that Gemma 2B has the largest vocabulary (256K tokens), followed by Llama3 (128K) and Mistral (32K tokens), resulting in improved fertility (2.36 vs 2.87 for Llama3/Mistral). Regardless, the larger models, Mistral 7B and Llama3 8B, vastly outperform the Gemma 2B model, suggesting fine-tuning smaller vocabulary LLMs like Mistral might be a better option from both cost and performance standpoints.

    \item \textbf{SFT alone is effective, but CPT+SFT yields best results.} While SFT yields large gains over prompting, combining CPT and SFT seems optimal for both Gemma and Llama3, although the gap is smaller than that of high-resource MT \citep{xu2024a}, ostensibly due to the difference in scale. For Mistral, SFT alone seems to suffice---we hypothesize that this might be due to the smaller vocabulary being effectively fine-tuned on SFT data alone.

    \item \textbf{Pre-training on parallel data yields major gains consistently.} Lastly, we observe that pre-training on a mixture of \emph{concatenated} monolingual and parallel data (\textit{`mono + parallel'}) yields statistically significant gains over converting both as monolingual data (ie. \textit{`all mono'}; refer Section \ref{sec:continued-pt}). This trend is consistent for all 3 LLMs, with larger gains for the more effective models Mistral and Llama3. We note that for `mono+parallel', we mix parallel and monolingual data in a 1:1 ratio, since we observed it worked best empirically, and show in Table \ref{tab:parallel_vs_mono_data_ratio} how increasing the ratio of parallel data during CPT monotonically improves performance. Given these gains, we explore the importance of bitext for low-resource LLM-MT further in Section \ref{sec:analysis-paralleldata}.
\end{enumerate}

\subsection{Analysis: Importance of Parallel Data} 
\label{sec:analysis-paralleldata}

\begin{figure}[t]
    \centering
    \includegraphics[width=0.9\columnwidth]{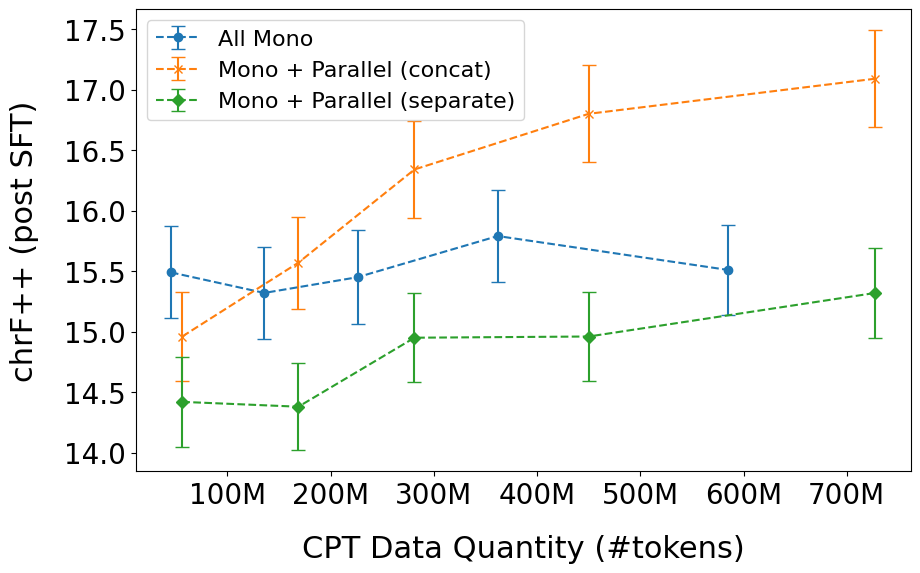}
    \caption{Comparing Llama3 8B models pre-trained on monolingual data alone versus those included parallel data too---concatenated, or as separate texts at various scales. All models were pre-trained on 1M, 3M, 5M, 8M, and 13M sentences respectively, and markers denote the corresponding token counts. The y-axis shows chrF++ post SFT on 500K spa-X MT data for 1 epoch.}
    \label{fig:cpt_tokens_vs_chrf}
\end{figure}

\paragraph{How important is concatenated parallel data at various scales of low-resource pre-training?} Having seen improvements in pre-training on the entire corpus (consisting of 13M `mono+parallel' sentences, or 730M tokens) in Table \ref{tab:primary-results}, we now study the importance of parallel data as we scale \emph{down}---an important consideration when one moves to even lower-resource settings. We pre-train on subsets of varying sizes and mix monolingual and parallel data in 3 ways: `All mono', `mono+parallel (concat)' and `mono+parallel (separate)', as defined in Section \ref{sec:continued-pt}). We fine-tune all these pre-trained models on the same SFT dataset: 500K spa-X MT instructions, and plot the resulting chrF++ scores in Figure \ref{fig:cpt_tokens_vs_chrf}, including error bars from bootstrap resampling. We note that `all mono' has different markers than the others, as the token counts on including both source and target-side data in the corpus are obviously larger than only the latter. 
We find that a) starting at around 5M sentences (\textasciitilde 300M tokens),  it is \emph{consistently advantageous} to include concatenated parallel data during pre-training. b) Given `mono+parallel (separate)' severely underperforms, we establish that it is \textbf{concatenation} that adapts the LLM for the task of MT, not the extra data alone. Our findings complement those of \citet{alves2024tower}, who also observe gains from pre-training on parallel data in the 1B to 20B tokens range, using 10 high-resourced European languages. In contrast, our focus here is on investigating the \textit{minimum data threshold} at which CPT leveraging parallel data becomes beneficial.

\begin{figure*}
    \resizebox{\textwidth}{!}{%
        \begin{minipage}{\textwidth}
            \centering
            \begin{subfigure}{0.32\textwidth}
                \includegraphics[width=\linewidth]{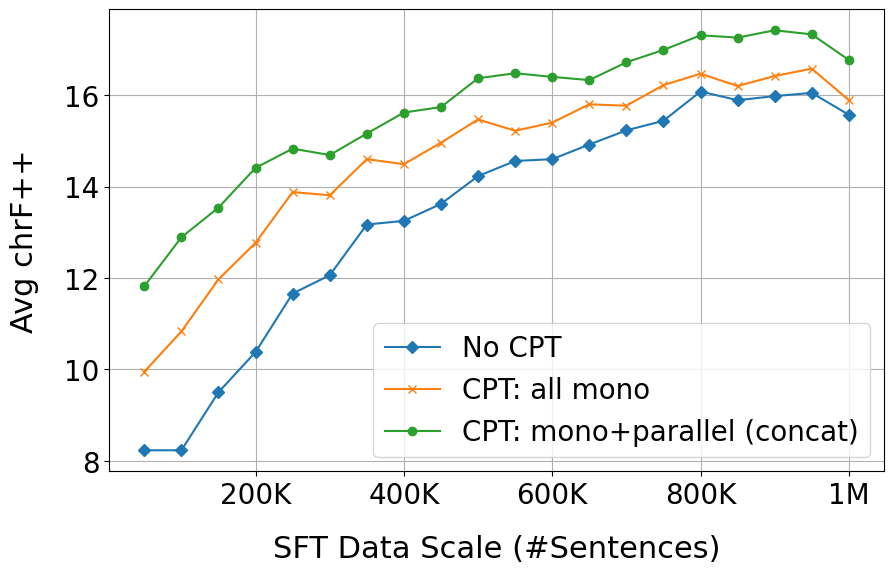} 
                \caption{Average (overall)}
                \label{fig:sftscale_overall}
            \end{subfigure}
            \hfill
            \begin{subfigure}{0.32\textwidth}
                \includegraphics[width=\linewidth]{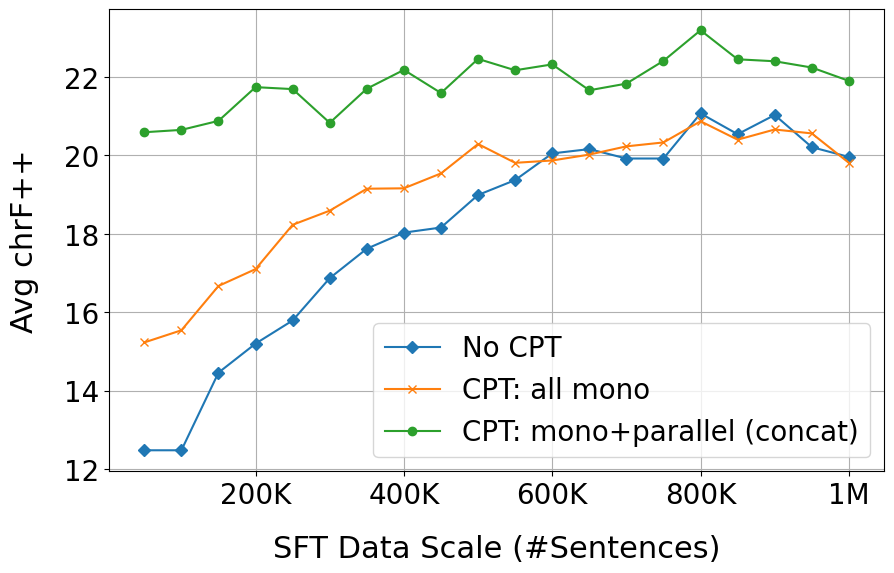} 
                \caption{Average (high-resource$^\dag$)}
                \label{fig:sftscale_hrl}
            \end{subfigure}
            \hfill
            \begin{subfigure}{0.32\textwidth}
                \includegraphics[width=\linewidth]{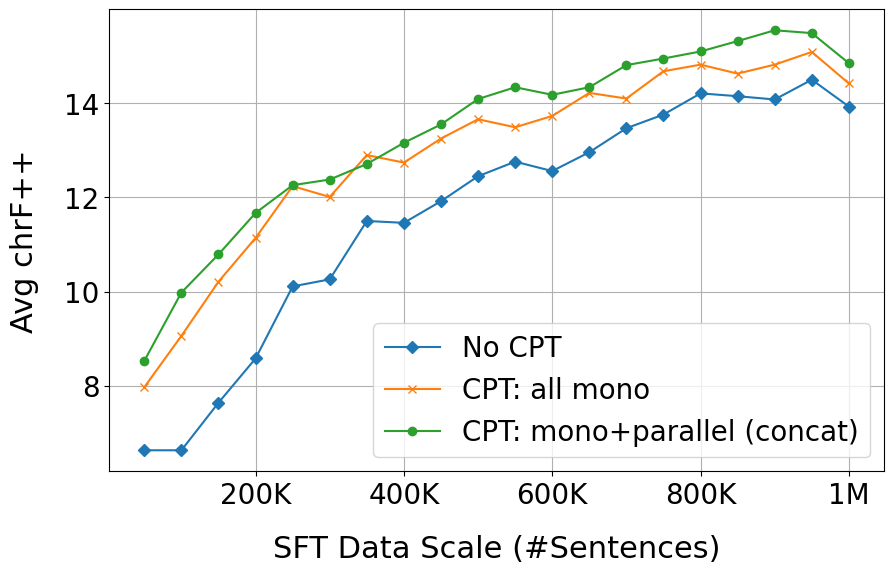} 
                \caption{Average (low-resource$^\eta$)}
                \label{fig:sftscale_lrl}
            \end{subfigure}
        \end{minipage}
    }
    \caption{Scaling up Llama3 8B models with different CPT recipes (no CPT, CPT with monolingual data, and CPT with a mixture of monolingual and parallel data) on MT data for the American languages. $^\dag$`High-resource' refers to the relatively higher-resourced languages in our low-resource setup (Aymara, Guarani and Quechua) while the other 8 are grouped as low-resource$^\eta$.}
    \label{fig:llama3-sft-scale}
\end{figure*}

\begin{figure}[t]
    \centering
    \includegraphics[width=0.75\columnwidth]{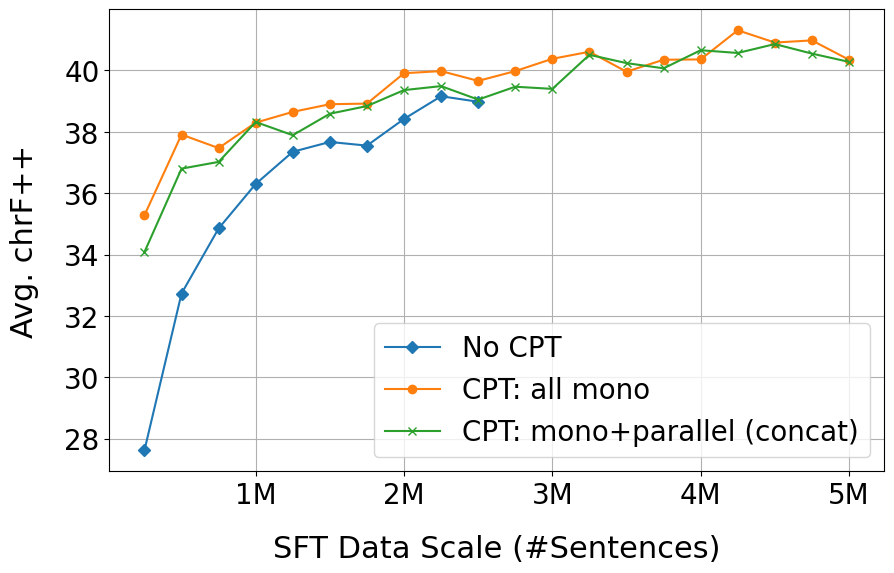}
    \vspace{-0.05in}
    \caption{Scaling up Llama 3 8B models with the 3 CPT recipes for the 4 Indic languages, until 5M sentences. We were forced to stop training `No CPT' at 2.5M sentences, constrained by budget.}
    \label{fig:indic_sft_scaling}
\end{figure}

\paragraph{How does scaling LRL parallel data during SFT impact performance?} We now turn our focus to scaling during SFT, which has not yielded gains in high-resource LLM-MT after 20K-30K sentences \citep{zhang2024when}.  Motivated by our previous results, we re-examine this question for low-resource LLM-MT in Figure \ref{fig:llama3-sft-scale}, wherein we evaluate at steps of 50K until \textasciitilde1M SFT instructions, the point at which spa-X MT data runs out. Our findings are:

\begin{enumerate}
    \item \textbf{Overall MT quality improves steadily with scale.} Unlike high-resource LLM-MT, we observe that, despite fluctuations, MT quality continues to grow until 1M sentences, particularly for LRLs, which have a steeper slope.

    \item \textbf{The gains of CPT on parallel data carry over to SFT scaling, particularly for HRLs.} 
    We observe that CPT on a mixture of concatenated monolingual and parallel data yields the largest initial gains, followed by CPT on monolingual-only data. We note that the former is particularly beneficial for HRLs, which have likely consumed a lot of parallel data during pre-training, and this helps them maintain a huge lead at lower SFT scales, suggesting that \textit{concatenated parallel data teaches the LLM the task of translation and helps it adapt to MT more naturally during SFT}. Given the prevalence of bitext in LLM pre-training corpora \citep{briakou-etal-2023-searching}, the gains of concatenation would help explain why few-shot prompting \citep{garcia2023unreasonable} and tiny-scale SFT (32 examples; c.f. \citet{zhu2024fine}) can elicit MT in the highest-resource languages!

    \item \textbf{LRLs need SFT scaling much more than HRLs} 
    So, while HRLs benefit hugely from CPT on bitext, the opposite is true for LRLs---the scarce amount of parallel data observed during CPT (see Table \ref{tab:american_parallel_stats}) is likely not enough to outperform `all mono' pre-training. Instead, LLMs are far more responsive to the scale of LRL MT data during SFT, showing consistent performance improvement.
    Meanwhile, MT quality plateaus for the HRLs\footnote{although Aymara, Guarani and Quechua \emph{are} LRLs for the original pre-training corpus of Mistral/Llama3} (Figure \ref{fig:sftscale_hrl}), suggesting that the `less is more' \citep{zhou2023lima} trend popular in high-resource LLM-MT does \emph{not} hold for LRLs, where \textit{scale continues to remain the most effective option.}

\end{enumerate}

For generalizability, we also provide scaling graphs for Mistral 7B in Figure \ref{fig:mistral7b-sft-scale} and report trends similar to Llama3. Moreover, we also compute scaling graphs for the 4 Indic languages in Figure \ref{fig:indic_sft_scaling} using the Llama3 model, until 5M eng-X sentences. Here, we observe that the gap between `CPT: mono+parallel (concat)' and `CPT (all mono)' is relatively lesser, and only significant until 500K sentences, at which point they start becoming comparable. This might be because the Indic parallel corpora used for CPT is relatively less diverse, consisting of only eng-X pairs in 4 languages, whereas our parallel data for the American languages is more heterogeneous (see Table \ref{tab:american_parallel_stats})---which we observed to yield more gains in our preliminary experiments. Thus, we expect a more diverse corpus for the Indic languages to yield greater long-term gains, but we leave the verification of this to future work. In other respects, the trends are quite similar to the American languages: the gains of CPT carry over to SFT here too and interestingly, MT quality continues to improve until 5M SFT examples, once again in direct contrast to high-resourced LLM-MT research.

\begin{figure}[t]
    \centering
    \includegraphics[width=0.75\columnwidth]{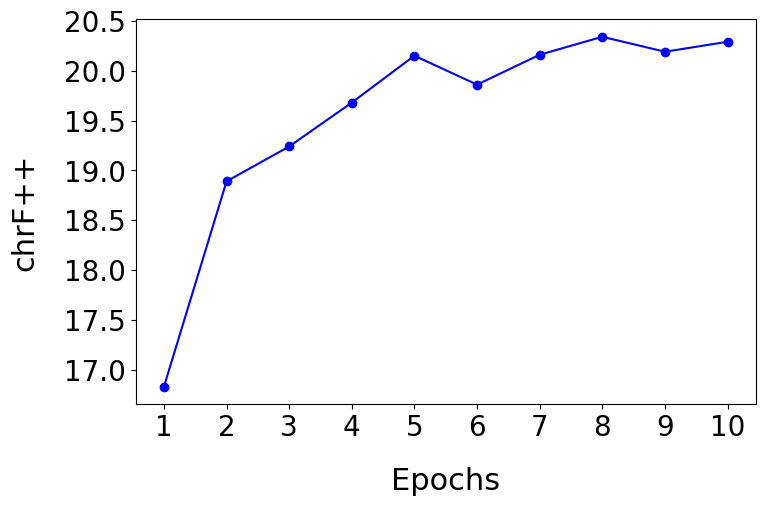}
    \vspace{-0.1in}
    \caption{Epoch vs performance graph for low-resource LLM-MT. We use the entire 1M spa-X MT dataset, and plot average chrF++ for the Indigenous American languages, using Llama3 (Mono+Parallel (concat)) model.}
    \label{fig:moreepochs}
\end{figure}

\paragraph{More epochs: Scale through repetition} Finally, given the consistent gains from scale, we evaluate how effective data repetition could be during SFT. We train for up to 10 epochs on the entire 1M spa-X MT dataset and plot our results in Figure \ref{fig:moreepochs}. We note a strong monotonic gain of +3.3 average chrF++ until 5 epochs, with the largest coming from the 1st to the 2nd epoch (+2.0 chrF++). While the graph does plateau later (likely due to overfitting) this shows how simply training for more epochs can be an easy way to boost performance, given the data constraints for LRLs.




\subsection{Analysis: Diversity in SFT is not \textit{always} helpful}
\label{sec:analysis-diversity}

We now look at the importance of \emph{diversity} in SFT, which plays a significant role in boosting LLM performance during general-purpose fine-tuning \citep{sanh2022multitask, pmlr-v202-longpre23a}, but has also been favoured in training LLM-MT systems \citep{alves2024tower}. \citet{alves2024tower} observed improved performance on non-MT tasks by including non-MT instructions, but mixed results on MT quality---concluding that the ``transfer/interference relations between tasks are complex''. In this work, we try to tease apart these interference relations and study the effect of non-MT tasks on low-resource translation in depth. In particular, we look at 3 kinds of diversity: \emph{prompt}, \emph{linguistic}, and \emph{task} diversity, and examine their impact on MT quality when mixed in varying proportions in the SFT dataset.

\begin{table}[t]
\centering\small
\setlength{\tabcolsep}{2ex}
\begin{tabular}{lccc}
\toprule
\textbf{Method} & \textbf{All} & \textbf{HRL} & \textbf{LRL} \\
\midrule
Bilingual LLM-MT & 14.51 & 24.55 & 10.75 \\
Multilingual LLM-MT & 18.73 & 23.59 & 16.90 \\
\bottomrule
\end{tabular}
\caption{Average chrF++ of SFT on a single language pair (bilingual) vs all language pairs (multilingual). As before, ``HRL'' includes Aymara, Guarani and Quechua, while the other eight are grouped under LRL.}
\label{tab:bilingual_vs_multilingual_sft}
\end{table}

\paragraph{Bilingual or Multilingual fine-tuning?} First, we ask whether given some multilingual low-resource data and a fixed FLOPS budget, would it be beneficial to do SFT on multiple, diverse low-resource MT pairs (multilingual LLM-MT) or only on a single pair (bilingual LLM-MT)? For this experiment, we simply concatenate all available data for the multilingual setting, while the bilingual setting consists of fine-tuning separate models for each pair, with the base LLM being constant in both cases: Llama3 (CPT with mono+parallel data). We show our results in Table \ref{tab:bilingual_vs_multilingual_sft}. We observe that, on average, multilingual SFT outperforms the bilingual models by +4 chrF++ points. Looking closer, we notice that these gains mostly come from the lower-resourced languages (detailed results in Table \ref{tab:bilingual-multilingual-llmmt}). This is potentially because the bilingual setting has too little data for effective FT in these languages, while the multilingual option offers better transfer and scale. For higher-resourced pairs, the opposite is true: the performance is slightly lower, ostensibly due to negative interference. This is not unlike conventional multilingual NMT models, such as NLLB, wherein high-resource MT can often be worse than bilingual baselines.

\begin{table}[t]
    \centering
    \setlength{\tabcolsep}{0.5ex}
    \begin{tabular}{@{}c@{}} 
\begin{subtable}[t]{\columnwidth} 
    \centering
    \small 
    \setlength{\tabcolsep}{0.3em} 
    \renewcommand{\arraystretch}{1.1} 
        \resizebox{0.99\linewidth}{!}{%
    \begin{tabular}{@{}l c@{}}
        \toprule
        \textbf{SFT Mixture} & \textbf{chrF++} \\
        \midrule
        \multicolumn{2}{c}{\textbf{Prompt Diversity}} \\
        \midrule
        500K spa-X MT (Same prompt) & 16.18\hspace{0.2ex}±\hspace{0.2ex}0.37 \\
        500K spa-X MT (Random prompts) & \textbf{17.22\hspace{0.2ex}±\hspace{0.2ex}0.40} \\
        \midrule
        \multicolumn{2}{c}{\textbf{Linguistic Diversity}} \\
        \midrule
        166K spa-X MT + 166K eng-X MT + 166K por-X MT & 14.26\hspace{0.2ex}±\hspace{0.2ex}0.37 \\
        250K spa-eng MT + 250K spa-X MT & 15.73\hspace{0.2ex}±\hspace{0.2ex}0.38 \\
        500K spa-X MT only & \textbf{17.22\hspace{0.2ex}±\hspace{0.2ex}0.40} \\
        \midrule
        \multicolumn{2}{c}{\textbf{Task Diversity}} \\
        \midrule
        250K spa-X MT + 250K XQA & 15.45\hspace{0.2ex}±\hspace{0.2ex}0.39 \\
        250K spa-X MT + 250K Aya (spa) & 15.68\hspace{0.2ex}±\hspace{0.2ex}0.39 \\
        250K spa-X MT (x 2 epochs) & 17.20\hspace{0.2ex}±\hspace{0.2ex}0.40 \\
        500K spa-X MT & \textbf{17.22\hspace{0.2ex}±\hspace{0.2ex}0.40} \\
        \bottomrule
    \end{tabular}%
    }
    \caption{Prompt, Linguistic and Task Diversity in American pairs}
    \vspace{0.1in}
    \label{tab:diversity_metrics}
\end{subtable} \\ 

\begin{subtable}[t]{\linewidth} 
    \centering
    \small 
    \setlength{\tabcolsep}{0.3em} 
    \resizebox{0.95\linewidth}{!}{%
    \begin{tabular}{@{}l@{\hspace{4em}}c@{}}

        \toprule
        \textbf{SFT Mixture} & \textbf{chrF++} \\
        \midrule
        250K spa-X MT + 250K Aya (asm, mni) & 26.91 ± 0.42 \\
        250K spa-X MT + 250K Alpaca (eng) & 28.02 ± 0.41 \\
        500K spa-X MT only & \textbf{29.57 ± 0.43} \\
        \bottomrule
    \end{tabular}
    }
    \caption{Task Diversity in Indic pairs}
    \label{tab:diversity_metrics_indic}
\end{subtable}
    \end{tabular}
        \caption{Exploring interference due to diversity in SFT for our best Llama3 8B model (CPT on both monolingual and parallel data) on the American and Indic languages. The dataset size (example count) is prepended before each task. Scores shown are average chrF++.}
        \vspace{-0.1in}
\end{table}
\paragraph{On Linguistic, Task and Prompt Diversity} Inspired by the previous results showing target-side diversity in MT pairs boosts performance, we now broaden the scope of diversity during SFT. In Table \ref{tab:diversity_metrics}, we show varying mixtures of SFT data that all have the same size (500K examples) but are composed of different tasks. For \emph{prompt diversity}, we ablate randomly sampling from a list of potential prompt templates (listed in Table \ref{tab:MTInstructionTemplates}) versus using a constant one (the first in Table \ref{tab:MTInstructionTemplates}). We observe a statistically significant gain in MT quality, similar to general-purpose SFT \citep{pmlr-v202-longpre23a}.

Next, we study the more interesting question of \emph{linguistic diversity}: can data in other MT pairs transfer for test languages? In one baseline, we divide our SFT dataset equally into spa-X, eng-X, and por-X MT examples, into the American languages (statistics in Table \ref{tab:american_parallel_stats}). In another baseline, we use Spanish-English data from ParaCrawl \citep{banon-etal-2020-paracrawl} and combine a 250K sentence subset with 250K pairs from our usual spa-X MT data. We find that this type of linguistic diversity leads to interference, and significantly underperforms the 500K spa-X MT baseline. This suggests that while \emph{target diversity} in related languages might help performance, source diversity or unrelated high-resource languages like English may not.

Then, we look at \emph{task diversity}: can non-MT tasks that elicit better instruction-following and general reasoning capabilities in the source or target language, benefit LLM-MT in LRLs? We mix Aya and XQA (Section \ref{sec:instruction-tuning}) instructions with a 250K subset of spa-X MT examples. We also include an ablation that, in place of curating non-MT data, merely trains for 2 epochs on the 250K spa-X MT subset. Here we also discover that general-purpose tasks lead to interference and that training for 2 epochs on 250K MT examples is a more effective strategy comparable to 1 epoch on 500K. 

We also look at the impact of task diversity on the Indic languages in Table \ref{tab:diversity_metrics_indic}. Here, we are able to find general-purpose instruction-tuning data in Assamese and Meitei, as mentioned in Section \ref{sec:instruction-tuning}. Interestingly, including data in these \emph{target languages}---the languages we want the model to be better at generating in---degrades performance the most! A similar amount of Alpaca data in English reduces performance far less, suggesting that reasoning in LRLs is such a hard task, that even data where the model \textit{should} learn to generate in the languages of interest, is \emph{worse} than generic English instruction-following data. Here, too, providing 500K spa-X examples is the most optimal strategy, underscoring the generalizability of our findings.

\begin{figure}[t]
    \centering
    \includegraphics[width=0.8\columnwidth]{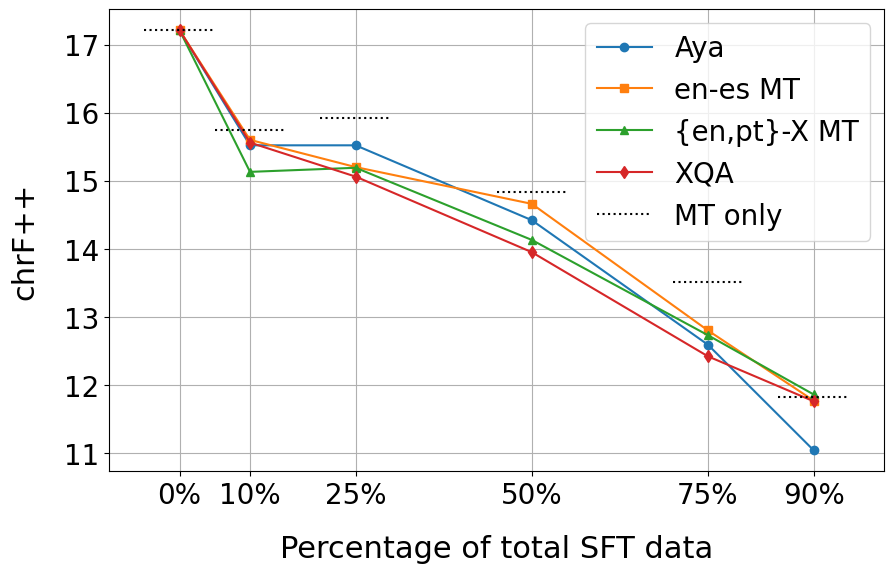}
    \caption{Mixing varying percentages of SFT tasks with spa-X MT, versus the impact on spa-X chrF++, for 500K examples. `MT only' is a topline which \emph{only} uses the spa-X MT data excluding all other tasks. Thus, it uses \emph{lesser}  data relatively and helps to estimate the interference of non-MT tasks on MT quality.}
    \vspace{-0.1in}
    \label{fig:diversity_tasks_ratio}
\end{figure}

\paragraph{Does non-MT SFT always cause interference?} 
Finally, we seek to establish conclusively if the negative results of diversity in Table \ref{tab:diversity_metrics} arise from factors like task mismatch or lack of quality; or if interference is simply \emph{the norm} when fine-tuning on non-MT data for low-resourced LLMs. In Figure \ref{fig:diversity_tasks_ratio}, we plot 5 graphs: the first 4 are graphs showing non-task-specific data (including XQA, Aya, \{eng/por\}-X MT and eng-spa MT) mixed in varying proportions with spa-X MT. Thus, for example, the second point of each plot represents 10\% of that task mixed with 90\% of spa-X MT, with the total always being 500K examples. We also include an ablation (dotted lines), which has the same quantity of spa-X MT data at each data point as other plots but without data from any other task listed. 

We observe that \textit{performance is always better using task-specific MT data}, and negative interference is indeed consistent. We conclude that transfer is challenging to achieve in low-resource settings, likely because at such scales, the LLM-MT model is still learning to generate translations, and reasoning is still a formidable challenge. Downstream performance in such settings depends not on the diversity/composition of the SFT dataset, but only on the amount of LRL MT data provided---making optimizing for \textit{quantity} the most effective strategy.


\subsection{Discussion: LLMs vs NMT models}

For the Indigenous American languages, the current SOTA systems are NordicAlps \citep{attieh2024system} and DC-DMV \citep{degenaro2024experiments} that report average chrF++ scores of 26.73 and 23.76 respectively. Both are NMT models. NordicAlps trained a multilingual model from scratch on varying mixtures of eng-spa and spa-x data at scale, with their most major gains (+4 chrF++ points) coming from a novel redundancy-driven tokenization method. \citet{degenaro2024experiments} fine-tune NLLB-200 using a variety of parallel data sources, but unlike us, they generate a large amount of synthetic data. Our best-performing model, Llama3 with CPT on parallel data and 5 epochs of SFT, yields a score of 20.93 in the best setting. While this is a gain of +17.06 chrF++ from a 5-shot prompted Llama3, it is also almost 6 chrF++ points behind the SOTA. Since ours was not a shared task effort, we did not attempt explorations with tokenization or synthetic data, which are concurrent to our findings and would likely boost performance further. Overall, our results suggest that low-resourced LLM-MT systems, while promising, are still behind the curve compared to SOTA NMT models.

\section{Conclusion}
\label{sec:conclusion}
In this paper, we approach low-resource LLM-MT from a data-centric perspective and study performance along two axes: i) the size of parallel data and ii) diversity in SFT tasks. Through experiments and analyses, we conclude that quantity plays a dominant role in downstream performance. Specifically, parallel data drives performance significantly during both CPT and SFT, with HRLs and LRLs displaying different behaviours---making bitext a critical resource even in the modern era of LLM-MT. Moreover, we establish that diversity (on multiple fronts) consistently declines MT quality, with multilingual fine-tuning on task-specific data being the most effective option, reaffirming our previous findings on the value of scale. We hope these findings will be useful considerations when scaling to massively multilingual LLMs of the future.

\section*{Limitations}
\label{sec:limitations}
One of the limitations of this work is that due to the unavailability of robust neural metrics for LRLs, we are forced to use string-based metrics like chrF++ as our primary evaluation metric. We note that while this is not optimal, it is also not unlike most other related works on low-resource translation, and chrF++ has been shown to align reasonably well with human assessment scores for morphologically rich languages. Secondly, we only focus on two specific language groups in this work. We expect that expanding the scope to a massively multilingual setting might yield even larger improvements, owing to scale and transfer learning.

\section*{Acknowledgments}
The authors have received funding from UK Research and Innovation (UKRI) under the UK government's Horizon Europe funding guarantee [grant numbers 10052546 and 10039436].

Computations described in this research were supported by the Baskerville Tier 2 HPC service (\url{https://www.baskerville.ac.uk}). Baskerville was funded by the EPSRC and UKRI through the World Class Labs scheme (EP/T022221/1) and the Digital Research Infrastructure programme (EP/W032244/1) and is operated by Advanced Research Computing at the University of Birmingham.

\bibliography{acl_latex}

\newpage
\appendix

\section{Data}
\label{sec:appendix}

\subsection{Monolingual Data}
\label{sec:appendix-cptdata}

We provide the detailed monolingual data statistics for the indigenous American languages used for CPT in Table \ref{tab:MonoStatsDetailed}. We collect this data from various sources. For the indigenous American languages, we use MADLAD-400 \citep{kudugunta2024madlad}, GLOT 500 \citep{imanigooghari-etal-2023-glot500}, Wikipedia, data curated by the University of Helsinki \citep{de-gibert-etal-2023-four} and OCR data collected by \citet{iyer-etal-2024-exploring}. The English and Spanish data used for replay comes from Wikipedia and MADLAD-400 too. We use MADLAD-400 and GLOT 500 again for the Indic languages again, along with the verified split of Sangraha \citep{khan2024indicllmsuite}---a large corpus for Indian languages.

\subsection{Parallel Data}

We provide the templates used for generating Machine Translation and XQA Instructions in Tables \ref{tab:MTInstructionTemplates} and \ref{tab:XQAInstructionTemplates} respectively. For MT, we randomly use one of these prompts to create an instruction for doing SFT of our model. For XQA, the provided instructions are used to prompt Mixtral-8x-7B to generate a question, which is later used for creating the synthetic XQA datasets as described in Section \ref{sec:instruction-tuning}.

We source the parallel data for generating these from the work of \citet{iyer-etal-2024-exploring}, which includes data sourced from AmericasNLP'23 official training datasets \citep{ebrahimi2023findings}, NLLB-200 and FLORES-200 \citep{costa2022no}, OPUS \citep{tiedemann2009news} and data curated by Helsinki in their 2023 submission \citep{de-gibert-etal-2023-four}. 

\subsubsection{Cleaning Parallel Data}

\paragraph{Rule-based Filtering} We clean parallel data by following standard filtering rules. First, we remove pairs with more digits or non-alphanumeric characters than alphabetic ones on either side. Then, we remove pairs where either the source or the target has less than 3 words or more than 120 words. To adjust for pairs containing partial or incomplete translations, we apply a combination of two filtering rules: a) for sentences with less than 25 words, the character length difference must be less than 65, and b) for those longer than 25 words, the character length ratio between either source and target, or vice versa, must be at most 1.55. Lastly, we filter our sentences with non-Latin characters in either sentence, as well as pairs with identical source and target sentences to create our final training dataset. 

\paragraph{Neural Quality Estimation} We also experimented with using neural quality estimation using models like LASER-3 \citep{heffernan-etal-2022-bitext}. LASER-3 supports 3 of the 11 American languages---the HRLs (Aymara, Guarani and Quechua)---and we used it for scoring parallel sentences for these languages. We then sorted them in order of decreasing quality, calculated as the cosine similarity of LASER representations. For the other 8 LRLs, we used random scoring to simulate standard shuffling behaviour. We found that training on this `sorted' data actually performed \textit{worse} than our default baseline trained on unscored corpora, consistently at various SFT scales. As none of the authors spoke these languages, we could not substantively explain why, but we hypothesize that LASER representations might not be very reliable for these very LRLs, and might introduce certain undesirable biases in the sorted data, making the default baseline more robust to different kinds of bitext. In practice, we found that the standard rule-based filtering approaches worked best and thus, we stuck to them for cleaning our data.

Finally, we note that both our monolingual and parallel corpora span a variety of domains similar to our test data. Also, due to the paucity of data, we use all available sources.
\begin{table*}[!t]
    \centering
    \resizebox{\textwidth}{!}{%
    \begin{tabular}{lcclcclcc}
        \toprule
        Language & \#Sentences & \#Tokens & Language & \#Sentences & \#Tokens & Language & \#Sentences & \#Tokens \\
        \midrule
        Aymara (\texttt{aym}) & 1M & 23.4M & Quechua (\texttt{quy}) & 1.9M & 45.1M & Guarani (\texttt{grn}) & 1.1M & 37.6M \\
        Nahuatl (\texttt{nhe}) & 1.1M & 32.1M & Otomi (\texttt{oto}) & 0.6M & 23.6M & English (\texttt{eng}) & 0.4M & 9.8M \\
        Spanish (\texttt{spa}) & 0.7M & 27.9M & Shipibo-Konibo (\texttt{shp}) & 0.1M & 3.3M & Bribri (\texttt{bzd}) & 0.1M & 2.6M \\
        Asháninka (\texttt{cni}) & 0.2M & 2.2M & Chatino (\texttt{ctp}) & 0.3M & 5.4M & Huichol (\texttt{hch}) & 0.2M & 3.5M \\
        Tarahumara (\texttt{tar}) & 0.2M & 2.3M & \textbf{Total (Replay)} & \textbf{1.1M} & \textbf{37.6M} & \textbf{Total} & \textbf{7.8M} & \textbf{218.9M} \\
        \bottomrule
    \end{tabular}
    }
    \caption{Detailed monolingual data statistics for the American languages}
\label{tab:MonoStatsDetailed}

\end{table*}
\begin{table*}[!h]
\centering
\begin{tabular}{@{}p{\linewidth}@{}}
\toprule
\textbf{Translation Instructions} \\ \midrule
1. Translate the following sentence from \texttt{\{src\_lang\}} to \texttt{\{tgt\_lang\}}. \\ 
2. Can you convert the following sentence from \texttt{\{src\_lang\}} to \texttt{\{tgt\_lang\}}. \\ 
3. Kindly translate this sentence from \texttt{\{src\_lang\}} into \texttt{\{tgt\_lang\}}. \\ 
4. Could you translate the following from \texttt{\{src\_lang\}} to \texttt{\{tgt\_lang\}}? \\ 
5. Proceed to translate the subsequent sentence from \texttt{\{src\_lang\}} to \texttt{\{tgt\_lang\}}. \\ 
6. Change the following sentence from \texttt{\{src\_lang\}} to \texttt{\{tgt\_lang\}}. \\ 
7. Render the sentence below from \texttt{\{src\_lang\}} into \texttt{\{tgt\_lang\}}. \\ 
8. Switch the following sentence from \texttt{\{src\_lang\}} into \texttt{\{tgt\_lang\}} language. \\ 
9. Rephrase the following sentence into \texttt{\{tgt\_lang\}} from \texttt{\{src\_lang\}}. \\ 
10. Transform the following text from \texttt{\{src\_lang\}} to \texttt{\{tgt\_lang\}}. \\ 
11. Can you restate the following sentence from \texttt{\{src\_lang\}} in \texttt{\{tgt\_lang\}}? \\ 
12. Please provide a translation for this sentence from \texttt{\{src\_lang\}} to \texttt{\{tgt\_lang\}}. \\ 
13. Adapt the following into \texttt{\{tgt\_lang\}} from the original \texttt{\{src\_lang\}}. \\ 
14. Translate the subsequent text from \texttt{\{src\_lang\}} into the \texttt{\{tgt\_lang\}} language. \\ \bottomrule
\end{tabular}
\caption{MT Instruction Templates used during Supervised Fine-Tuning (SFT)}
\label{tab:MTInstructionTemplates}
\end{table*}

\begin{table*}[!h]
\centering
\begin{tabular}{@{}p{\linewidth}@{}}
\toprule
\textbf{XQA Instruction} \\ \midrule
"Consider this sentence: \texttt{\{input\}}\textbackslash nWhat kind of specific instruction X could this be the unique answer to? Output ONLY the instruction, followed by a newline." \\ \addlinespace \bottomrule
\end{tabular}
\caption{Template used for generating XQA instructions}
\label{tab:XQAInstructionTemplates}
\end{table*}

\begin{table*}[!h]
\centering
\setlength{\tabcolsep}{3pt} 
\resizebox{\textwidth}{!}{%
\begin{tabular}{@{}lcccccccccccc@{}}
\toprule
Method & Avg & es-aym & es-bzd & es-cni & es-ctp & es-gn & es-hch & es-nhe & es-oto & es-quy & es-shp & es-tar \\
\midrule
Bilingual LLM-MT & 14.5 & \textbf{21.2} & 6.1 & 12.3 & 8.4 & \textbf{27.9} & 15.0 & \textbf{14.1} & 9.5 & 24.6 & 14.5 & 6.1 \\
Multilingual LLM-MT & \textbf{18.7} & 20.2 & \textbf{15.8} & \textbf{18.0} & \textbf{26.0} & 25.8 & \textbf{21.6} & 18.2 & \textbf{11.2} & \textbf{24.9} & \textbf{14.8} & \textbf{9.5} \\
\bottomrule
\end{tabular}%
}
\caption{Comparison of Bilingual and Multilingual Es-X FT (MT only) Methods}
\label{tab:bilingual-multilingual-llmmt}
\end{table*}

\section{Results}
\begin{table}[!h]
\centering
\begin{tabular}{l@{\hskip 0.5cm}c@{\hskip 0.5cm}c@{\hskip 0.5cm}c}
\toprule
\textbf{Ratio} & \textbf{Gemma 2B} & \textbf{Mistral 7B} & \textbf{Llama3 8B} \\
\midrule
0\% & 9.6 ± 0.3 & 15.6 ± 0.4 & 15.6 ± 0.4 \\
10\% & 10.1 ± 0.3 & 15.9 ± 0.4 & 16.5 ± 0.4 \\
25\% & 10.1 ± 0.3 & 16.6 ± 0.4 & 16.6 ± 0.4 \\
50\% & 10.1 ± 0.3 & 16.7 ± 0.4 & 17.2 ± 0.4 \\
\bottomrule
\end{tabular}
\caption{Model Performance vs. Parallel Data Ratio. While the gap between ratios is not always statistically significant, it is clear that the trend is monotonic and having 50\% parallel data is consistently better than 0\% (ie. the fully monolingual setting).}
\label{tab:parallel_vs_mono_data_ratio}
\end{table}
\subsection{Parallel vs Monolingual Data Ratio}

In Table \ref{tab:parallel_vs_mono_data_ratio}, we find that mixing higher ratios of parallel data with monolingual data either performs comparably or improves performance. We do not go higher than 50\% since our parallel data runs out at this stage, and to ensure higher ratios we would have to oversample the existing dataset---which would not lead to a fair comparison with the other baselines. Nevertheless, given the monotonic trend, it would be interesting to explore if mixing higher ratios of parallel data continues to improve the performance even more. However, it is likely that there is some ceiling as to how much parallel data one should mix. For instance, \citet{alves2024tower} show how a baseline trained on 100\% parallel data underperforms compared to mixing it with monolingual data.

\subsection{Scaling up SFT: Mistral 7B}
\begin{figure*}[!t]
    \resizebox{\textwidth}{!}{%
        \begin{minipage}{\textwidth}
            \centering
            \begin{subfigure}{0.32\textwidth}
                \includegraphics[width=\linewidth]{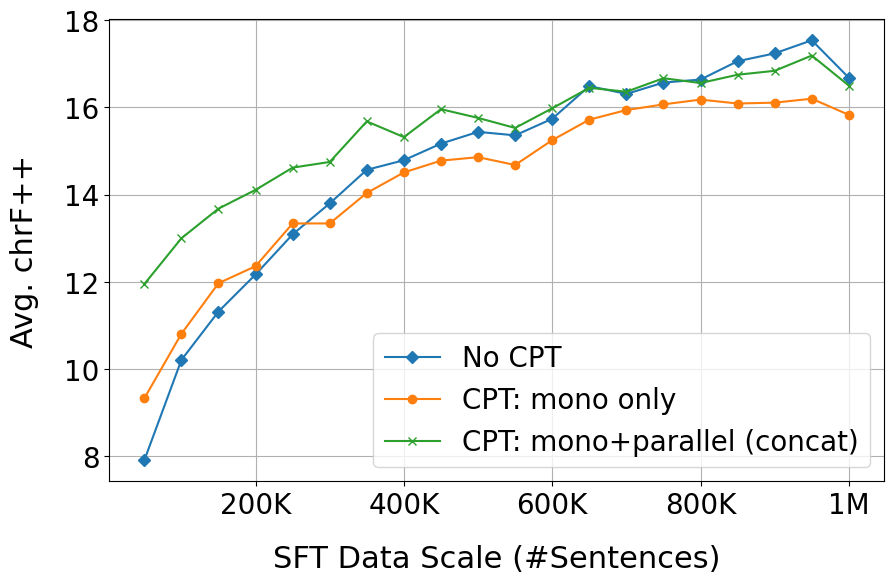} 
                \caption{Average (overall)}
                \label{fig:sftscale_overall_mistral}
            \end{subfigure}
            \hfill
            \begin{subfigure}{0.32\textwidth}
                \includegraphics[width=\linewidth]{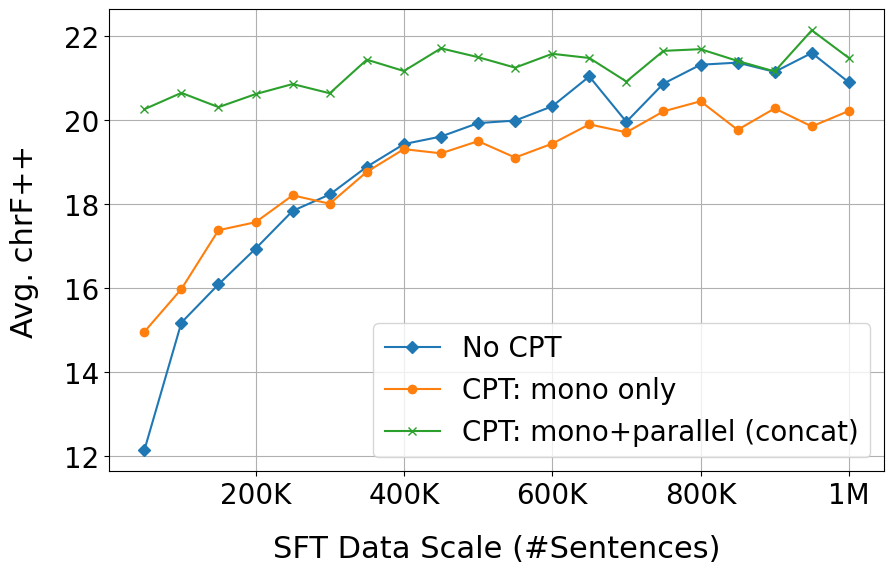} 
                \caption{Average (high-resource$^\dag$)}
                \label{fig:sftscale_hrl_mistral}
            \end{subfigure}
            \hfill
            \begin{subfigure}{0.32\textwidth}
                \includegraphics[width=\linewidth]{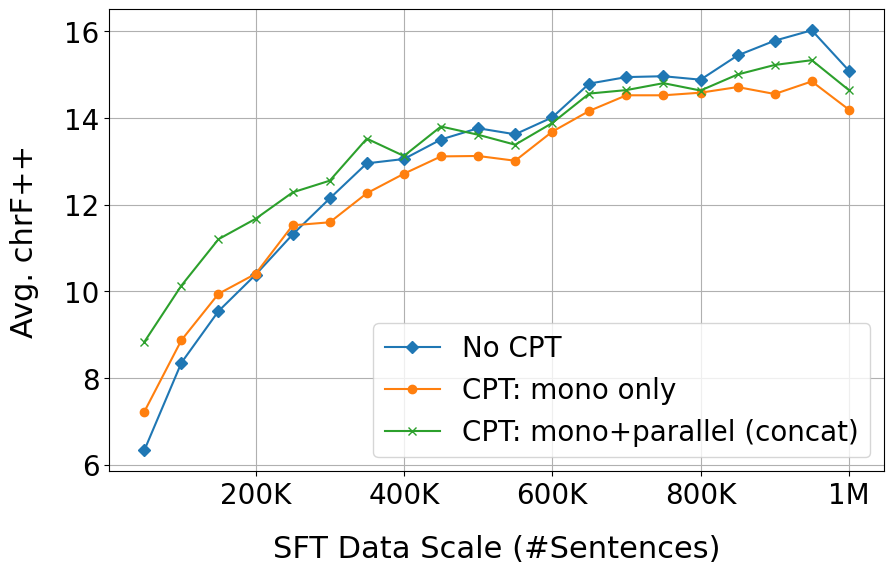} 
                \caption{Average (low-resource$^\eta$)}
                \label{fig:sftscale_lrl_mistral}
            \end{subfigure}
        \end{minipage}
    }
    \caption{Scaling up Mistral 7B models with different CPT recipes (no CPT, CPT with monolingual data, and CPT with a mixture of monolingual and parallel data) on MT data for the American languages. We observe that the trends are very similar to Llama3, with the sole exception that the gains of CPT diminish faster at around 500K sentences. This might be because Mistral, due to a 4x smaller vocabulary, gets fine-tuned effectively with less data. $^\dag$`High-resource' refers to the relatively higher-resourced languages in our low-resource setup (Aymara, Guarani and Quechua) while the other 8 are grouped as low-resource$^\eta$. }
    \label{fig:mistral7b-sft-scale}
\end{figure*}

Figure \ref{fig:mistral7b-sft-scale} shows the scaling behaviour of the Mistral 7B model with different pre-training recipes (no CPT, CPT with monolingual data, and CPT with a mixture of monolingual and parallel data), where terminology is the same as that defined in Section \ref{sec:continued-pt}.  We observe that the trends for Mistral are largely similar to Llama3:

\begin{enumerate}
    \item `No CPT' underperforms the CPT baseline, but they become comparable at about \textasciitilde500K sentences. As hypothesised in Section \ref{sec:foundational-results}, due to a 4x larger vocabulary, Llama3 has 4x more parameters to fine-tune\footnote{LoRA module parameters are negligible in comparison. For fine-tuning input and output embeddings, Llama3 has 128K * 4096 * 2 $\approx$ 1B parameters, whereas Mistral has 32K * 4096 * 2 $\approx$ 250M parameters. LoRA parameters are on the order of 3M, for comparison.}, meaning our Llama3 models are likely more data hungry and retain the benefits of CPT over longer periods, but smaller models overfit sooner, leading to a shorter `cross-over' threshold.
    
    \item `CPT (mono only)' consistently underperforms `CPT (mono + parallel)', very similar to Llama3, lending further credence to our conclusion that concatenated parallel data adapts the model to MT in a much better way

    \item For HRLs, the performance plateaus quite quickly, while for LRLs SFT quality continues to grow with scale, once again following Llama3's trend
\end{enumerate}

Our experiments with Mistral thus help in providing more evidence to support our claims with regard to the benefit of parallel data for low-resource LLM-MT systems.

\subsection{Bilingual vs Multilingual LLM-MT: Detailed Results}

We provide detailed, language-specific results for bilingual vs multilingual LLM-MT in Table \ref{tab:bilingual-multilingual-llmmt}. We observe that multilingual LLM-MT mostly outperforms bilingual baselines with the exception of 3 relatively higher-resourced languages (Aymara, Nahuatl and Guarani) where there is a bit of a gap.

\end{document}